\documentclass[10pt,twocolumn,letterpaper]{article}

\usepackage[pagenumbers]{iccv} %

\usepackage{mathtools}
\usepackage{nicefrac}
\usepackage{multirow}
\usepackage{overpic}
\usepackage{algorithm}
\usepackage{algpseudocode}

\newif\ifdraft
\draftfalse

\ifdraft

\newcommand{\TODO}[1]{\textbf{\color{red}[TODO: #1]}}
\newcommand{\dale}[1]{{\color{blue}[Dale: #1]}}
\newcommand{\mg}[1]{{\color{Orange}[MG: #1]}}
\newcommand{\vk}[1]{{\color{green}[VK: #1]}}

\newcommand{\rana}[1]{{\color{purple}[Rana: #1]}}
\newcommand{\yifan}[1]{{\color{teal}[Yifan: #1]}}

\else

\newcommand{\TODO}[1]{}
\newcommand{\dale}[1]{}
\newcommand{\mg}[1]{}
\newcommand{\vk}[1]{}

\newcommand{\rana}[1]{}
\newcommand{\yifan}[1]{}

\fi

\definecolor{rcyan}{rgb}{0.0,0.6,0.7}
\definecolor{rmagenta}{rgb}{0.7,0.0,0.6}

\usepackage{overpic}
\definecolor{iccvblue}{rgb}{0.21,0.49,0.74}
\usepackage[pagebackref,breaklinks,colorlinks,allcolors=iccvblue]{hyperref}
\usepackage{comment}

\def\paperID{66} %
\def\confName{ICCV}
\def\confYear{2025}

\title{Reusing Computation in Text-to-Image Diffusion \\
for Efficient Generation of Image Sets}

\author{Dale Decatur$^1$ \hspace{10mm}
Thibault Groueix$^2$ \hspace{10mm}
Wang Yifan$^2$ \hspace{10mm}
Rana Hanocka$^1$ \\
Vladimir Kim$^2$\hspace{10mm}
Matheus Gadelha$^2$
\\ \\
$^1$University of Chicago
\hspace{20mm}
$^2$Adobe Research
\vspace{2mm}
}

\begin{document}
\twocolumn[{%
\renewcommand\twocolumn[1][]{#1}%
\maketitle
\begin{center}
    \centering
    \newcommand{\pl}{38}
    \vspace{-3mm}
    \begin{overpic}[width=\textwidth]{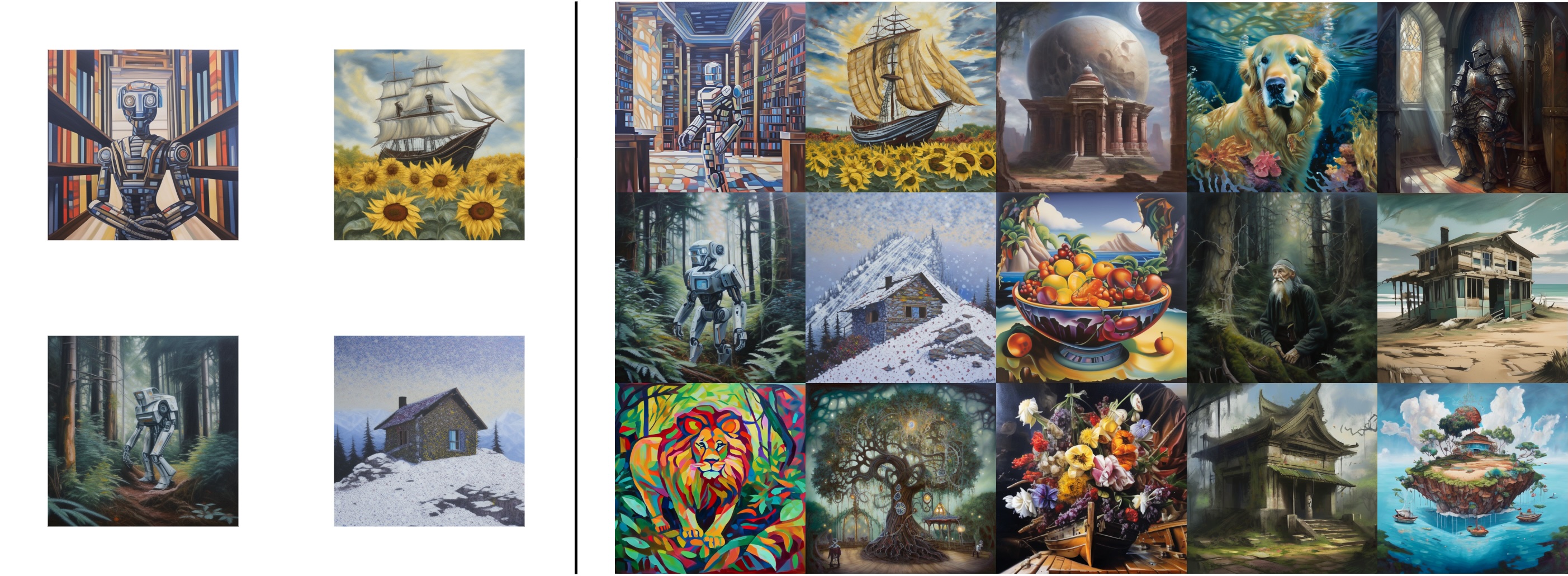}
    \put(11,  \pl){\textcolor{black}{Standard Diffusion}}
    \put(65,  \pl){\textcolor{black}{Our Method}}
    \end{overpic}
    \vspace{-7mm}
    \captionof{figure}{
    When generating a collection of images from text prompts, our approach shares denoising steps and can produce images of comparable or better quality while using a fraction of the compute. Here, we show images generated by the standard approach vs our approach, for a fixed compute budget.
    }
    \label{fig:teaser}
    \vspace{-1mm}
\end{center}

}]

\begin{abstract}
Text-to-image diffusion models enable high-quality image generation but are computationally expensive. While prior work optimizes per-inference efficiency, we explore an orthogonal approach: reducing redundancy across correlated prompts. Our method leverages the coarse-to-fine nature of diffusion models, where early denoising steps capture shared structures among similar prompts.
We propose a training-free approach that clusters prompts based on semantic similarity and shares computation in early diffusion steps. Experiments show that for models trained conditioned on image embeddings, our approach significantly reduces compute cost while improving image quality.
By leveraging UnClip’s text-to-image prior, we enhance diffusion step allocation for greater efficiency.
Our method seamlessly integrates with existing pipelines, scales with prompt sets, and reduces the environmental and financial burden of large-scale text-to-image generation.
\end{abstract}

\section{Introduction}
Diffusion models have revolutionized creative workflows, profoundly reshaping the landscape of image generation since their introduction~\cite{rombach2022high,lin2024sdxl,esser2024scaling,saharia2022photorealistic,razzhigaev2023kandinsky,kandinsky3}. Their impressive capability to generate realistic, high-quality images from textual descriptions has made them integral to both consumer applications and professional creative processes~\cite{firefly,dalle,runway,midjourney}. However, the significant computational cost associated with diffusion models poses a substantial barrier to their widespread practical deployment—generating a single image can consume energy equivalent to charging half a smartphone battery~\cite{luccioni2024power}.

In practical scenarios, users rarely generate just a single image. Instead, these models serve primarily as tools to visually explore ideas, leading users to produce and evaluate numerous image variations to refine and select the most satisfactory outputs. Community-driven methods, such as prompt templates and automatic prompt variation techniques~\cite{promptify,prompt4discovery,promptexpansion}, have emerged precisely to facilitate this extensive exploration of prompt variations. Consequently, the typical workflow for text-to-image generation inherently involves creating extensive image collections to effectively explore the ideation space encoded in textual inputs. Yet, despite the ubiquity of this approach, little research has explicitly targeted optimization across multiple prompts.

\begin{figure}
    \centering
    \newcommand{\pl}{-4}
    \begin{overpic}[width=\linewidth]{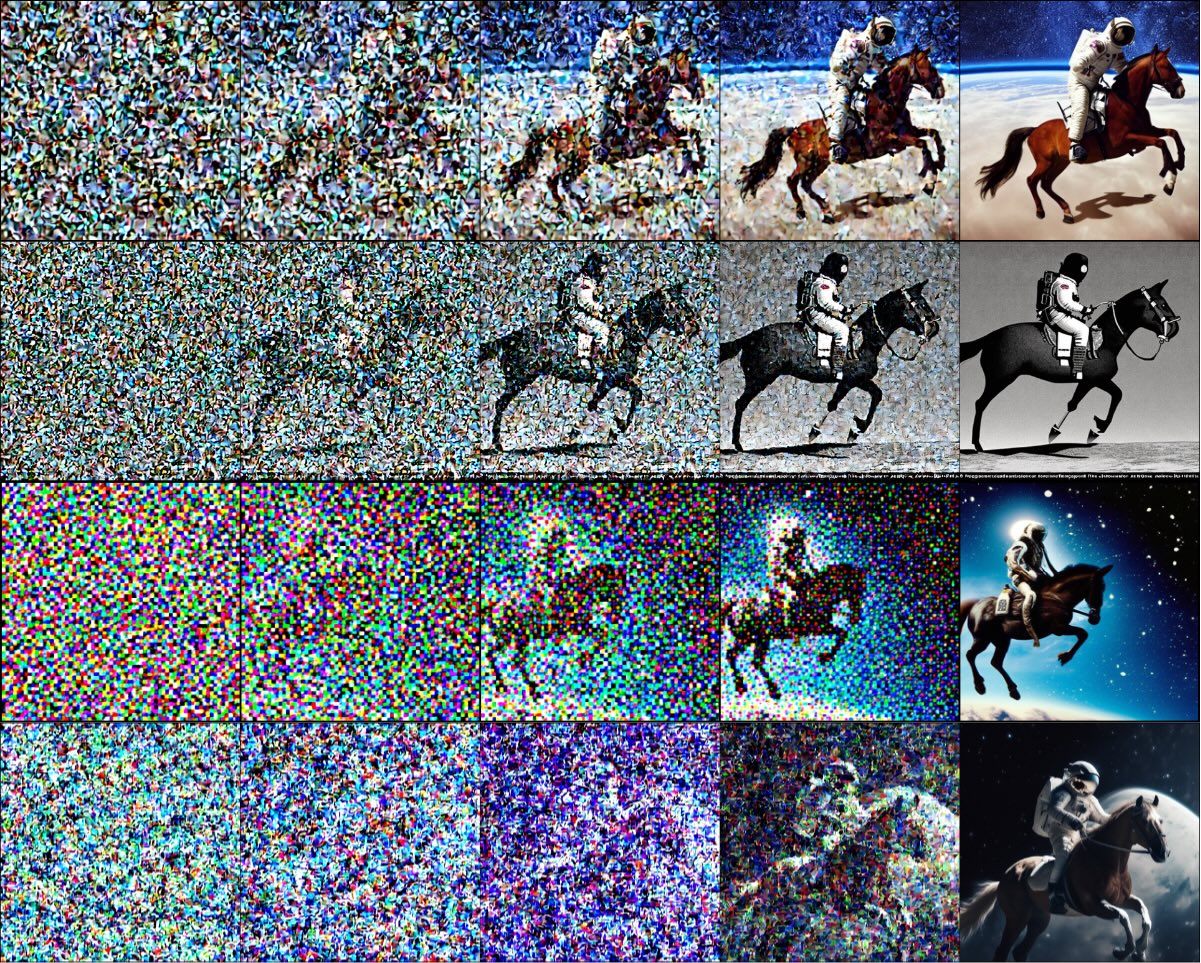}
    \put(4,  \pl){\textcolor{black}{Step 0}}
    \put(23,  \pl){\textcolor{black}{Step 10}}
    \put(43,  \pl){\textcolor{black}{Step 20}}
    \put(64,  \pl){\textcolor{black}{Step 30}}
    \put(85,  \pl){\textcolor{black}{Step 40}}
    \end{overpic}
    \vspace{-2mm}
    \caption{\textbf{Coarse-to-Fine Generation}. We show intermediate latents from the diffusion process for Stable Diffusion 1.5~\cite{rombach2022high} (top row), Stable UnCLIP~\cite{ramesh2022hierarchical} (second row), Karlo~\cite{kakaobrain2022karlo-v1-alpha} (third row), and Kandinsky~\cite{razzhigaev2023kandinsky} (bottom row). The models trained without a text-to-image prior (Stable Diffusion and Stable UnCLIP) learn structural details and high frequency features earlier on in the diffusion process. In contrast, Karlo and Kandinsky which were trained with the text-to-image prior, learn structural details later in  denoising  and are able to quickly add high frequency details in few steps, making them ideal for our compute sharing approach.
    }
    \vspace{-4mm}
    \label{fig:coarse-to-fine}
\end{figure}
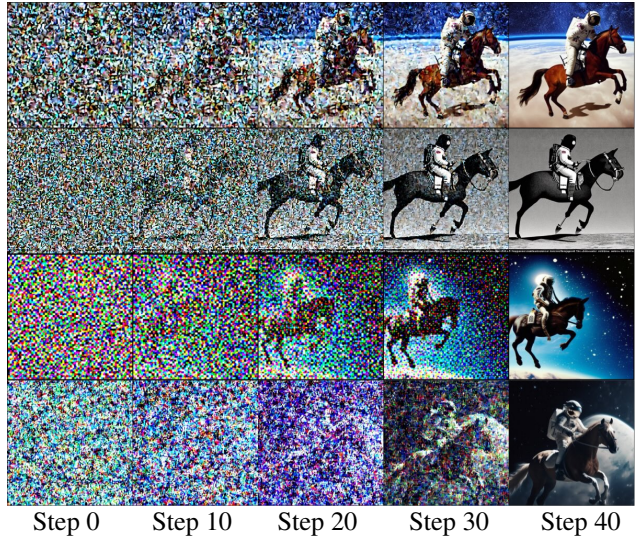

In this paper, we address this critical gap by proposing a novel method specifically designed to reduce computational overhead when performing large-scale inference on diffusion-based text-to-image models. While existing literature has primarily focused on per-inference cost optimization through methods such as distillation~\cite{yin2024one,luo2023latent}, we explore an orthogonal direction. Our core insight leverages the widely recognized property of diffusion models (see \cref{fig:coarse-to-fine}): they generate images in a coarse-to-fine manner~\cite{dalle,saharia2022photorealistic,rombach2022high}, where early denoising steps primarily define low-frequency, structural content (overall composition and layout), and subsequent steps progressively refine high-frequency, fine-grained details. Given the inherent statistical correlation among natural images at low frequencies, significant redundancy arises during initial diffusion steps for prompts describing visually distinct but structurally similar images -- for example, ``a Labrador wearing a bow-tie" and ``a Portuguese water dog wearing sunglasses".

Motivated by this observation, we explore the fundamental question: can we reuse early-stage computations across correlated prompts to increase efficiency? Through extensive experiments, we found that while all diffusion models generate images from coarse to fine, certain families exhibit a more evenly distributed progression, making them particularly suitable for early-stage computation sharing. For instance, as shown in \cref{fig:coarse-to-fine}, Stable Diffusion (SD)~\cite{rombach2022high} and Stable UnCLIP~\cite{ramesh2022hierarchical} generate fine details relatively early, limiting opportunities for shared computation. In contrast, models such as Kandinsky~\cite{razzhigaev2023kandinsky,kandinsky3} and Karlo~\cite{kakaobrain2022karlo-v1-alpha} exhibit a more gradual emergence of high-frequency details, allowing more efficient reuse of early-stage computations. While we hypothesize that training with a text-to-image prior may encourage this behavior, our method remains effective as long as the model exhibits a sufficiently gradual detail emergence, regardless of embedding strategy.

To put these findings into practice, we introduce a straightforward yet effective approach that utilizes off-the-shelf text encoders and agglomerative clustering to automatically construct hierarchical representations of prompts.
Specifically, given a set of prompts, we construct a hierarchical embedding tree, where similar concepts share their averaged embedding, used to guide early steps of the denoising process.
By exploiting this tree structure during diffusion, our method shares computations among prompts with structural similarities in early diffusion steps and progressively specializes the process as denoising advances.

Through extensive experimentation across diverse image collections and prompt variations -- including systematically generated prompts using established templates -- we demonstrate that our hierarchical diffusion procedure significantly reduces computational overhead, achieving comparable or even superior visual quality with as few as $26\%$ of the total diffusion steps required by conventional methods.

Our approach is entirely automatic, training-free, easily integrated into existing denoising pipelines, and scales favorably with the number of prompts. Crucially, our findings not only demonstrate substantial computational savings but also highlight unique generation dynamics of different diffusion models, contributing valuable insights into diffusion behavior. We hope our research will inspire further exploration into efficient and sustainable multi-prompt optimization strategies, ultimately reducing the substantial carbon footprint associated with diffusion models.

\section{Related Work}

\paragraph{Fast Image Generation.}

Image generation has evolved through two major breakthroughs: Generative Adversarial Networks (GANs)~\cite{goodfellow2014generative} and diffusion models~\cite{ho2020denoising, sohl2015deep}. GANs enable fast inference but suffer from training instability and poor generalization~\cite{gigagan, biggan}. Diffusion models, while more stable and expressive, require significantly more compute due to the iterative denoising at inference time. Methods such as latent-space diffusion~\cite{rombach2022high} mitigate some inefficiencies but do not fundamentally reduce the sampling burden.

To accelerate diffusion sampling, prior works propose efficient numerical solvers~\cite{karras2022elucidating, liu2022pseudo, lu2022dpm, lu2022dpm2, song2020denoising, zhao2023unipc}, reducing steps from 1000 steps~\cite{sohl2015deep, ho2020denoising} to 10~\cite{song2020denoising, karras2022elucidating, zhao2023unipc}, though in practice most models are sampled with about 40 steps to trade off between speed and quality.
Distillation-based methods~\cite{salimans2022progressive, lin2024sdxl, yin2025improved, yin2024one, kim2023consistency, luo2023latent} train models to generate images in fewer steps. While effective, these approaches require additional non-trivial training.
Complementary to these methods, our work addresses the common scenario of multiple prompts by providing a training-free acceleration approach that leverages hierarchical prompting to enable shared computations across related prompts.

\smallskip
\noindent\textbf{Prompt Expansion.}
Prompt expansion techniques help users explore diverse image variations by generating structured prompt sets. Large language models assist in expanding prompts while preserving semantic relevance~\cite{promptexpansion}. Community-driven prompt templates provide predefined tag structures~\cite{promptartists, mahdavi2024ai}, allowing systematic prompt variation.

Systems like DreamSheets~\cite{prompt4discovery} use spreadsheets to dynamically populate prompts, while Promptify~\cite{promptify} enhances user interactions with automated prompt suggestions. These tools enable efficient idea exploration. Our method aligns with this paradigm but focuses on efficient image generation when prompts are predefined, optimizing computation for structured workflows.

\smallskip
\noindent\textbf{Coarse-to-Fine Generation in Diffusion Models.}
Diffusion models generate images progressively, with lower-frequency details emerging in early denoising steps and high-frequency details refined later~\cite{ho2020denoising}. While this property exists across architectures, models trained with UnCLIP-style image embeddings~\cite{ramesh2022hierarchical} exhibit it more distinctly.

UnCLIP models introduce a text-to-image embedding prior, mapping CLIP text embeddings to image embeddings. When trained with UnCLIP image embeddings, the model exhibits improved diversity~\cite{ramesh2022hierarchical} and, as shown by Karlo~\cite{kakaobrain2022karlo-v1-alpha}, generates fine-grained details with fewer steps, which consequently affects the allocation of diffusion steps, allowing fine details to emerge later in the process~\cite{razzhigaev2023kandinsky,kandinsky3}.
Our method leverages this property effectively. It shares computation during early steps that generate low-frequency features, while dedicating fewer unique steps to produce the high-frequency details specific to each prompt.

\newcommand{\x}{\mathbf x}
\newcommand{\X}{\mathcal{X}}
\newcommand{\y}{\mathbf y}
\newcommand{\m}{\mathbf m}
\newcommand{\argmin}{\mathop{\operatorname{argmin}}\limits}

\setlength{\abovedisplayskip}{3pt}
\setlength{\belowdisplayskip}{3pt}

\section{Preliminaries: Text-to-Image Diffusion}
\noindent\textbf{Training.}
Given an input image $\x$ and text embedding $y$, a noisy image $\x_t$ is first obtained by adding a random amount of noise to $\x$:
\begin{equation}
    \label{eq:noising}
    \x_t = \sqrt{\alpha(t)}\ \x + \sqrt{1-\alpha(t)}\ \epsilon,
\end{equation}
where $\epsilon \sim \mathcal{N}(\mathbf{0}, \mathbf{I})$ is Gaussian noise, and $t \in [0, T]$ parameterizes a schedule $\alpha$ corresponding to the amount of noise in $\x_t$; \ie $\alpha(0) = 1$ (no noise) and $\alpha(T) = 0$ (pure noise).
$\epsilon_{\theta}$ is the denoiser -- a model trained
to invert this noising process by minimizing the following loss through gradient-descent: 
\begin{equation*}
    \label{eq:diffloss}
    \mathcal{L_{\text{diff}}} := w(t) \|\epsilon_\theta(\x_t; t, y) - \epsilon\|_2^2,
\end{equation*}
where
$w(t)$ is a weighting scheme parametrized by $t$.
In practice, all our experiments use latent diffusion models which means that the previous loss is computed in latent space and a VAE is used to decode it into pixels~\cite{rombach2022high}.

\smallskip
\noindent\textbf{Inference.}
Once $\epsilon_\theta$ is trained, a new image is generated using the following procedure.
Given a text embedding $y$, we start from pure noise $\x_T$ and iteratively denoise it:%
\begin{align}
    \x_T &\sim \mathcal N(\mathbf 0, \mathbf I) \nonumber \\
    \x_{t-1} &\sim \mathcal{N}(a_t \x_t - b_t \epsilon_\theta(\x_t; t, y), \sigma^2_t\mathbf{I})
    \label{eq:inference_step}
\end{align}
where $a_t$, $b_t$, the variance $\sigma^2_t$, and the timesteps are chosen according to a denoising schedule.
The number of steps taken ($K$) is a parameter of the the denoising scheduler.
$K$ has a strong correlation with the quality of the generated image.
After \cref{eq:inference_step} is applied $K$ times, $\x_t$ should yield
a fully denoised result.
The standard diffusion process generates a set of $N$ images by running \cref{eq:inference_step} independently for each text embedding, requiring $KN$ evaluations of $\epsilon_\theta(\x_t; t, y)$.

\begin{figure*}[t]
    \centering
    \includegraphics[width=\linewidth]{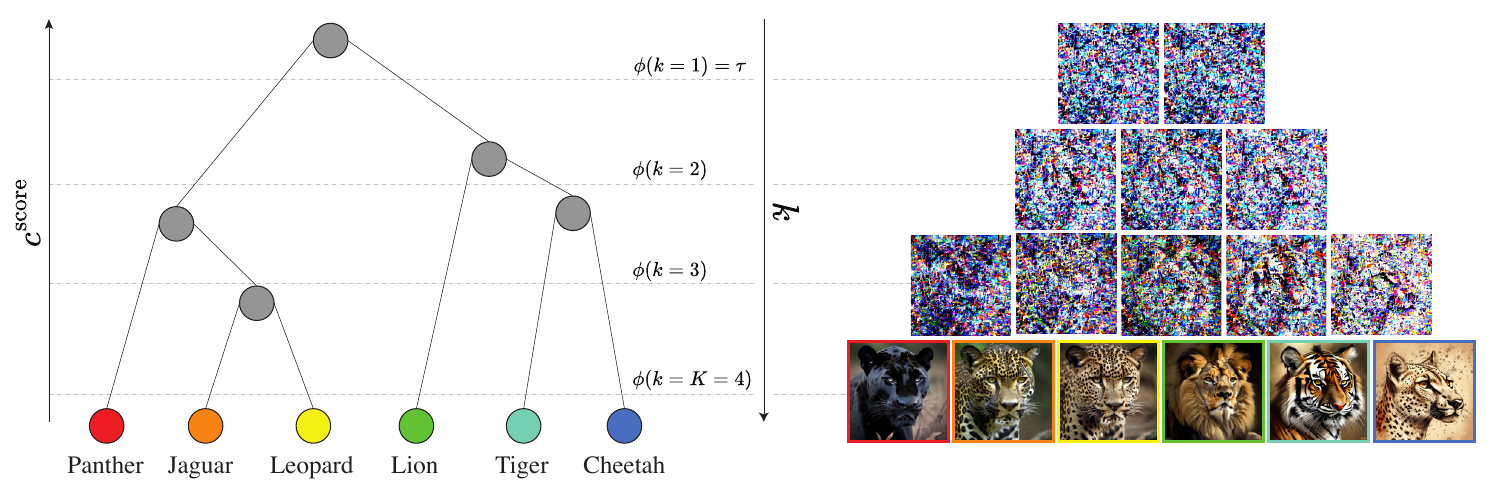}
    \vspace*{-15pt}
    \caption{\textbf{Tree Traversal}. \textit{Left}: our approach relies on a tree structure obtained by running agglomerative clustering on a set of prompt embeddings. Each node in the tree contains the average of the embeddings of its children, and a heterogeneity score $c^{\text{score}}$ based on the distance between its two children. To connect the tree hierarchy to the denoising steps, we design a function $\phi$ taking as input the denoising step $k$, and controlling via its output and $c^{\text{score}}$   which level to use in the tree for step $k$. \textit{Right}: As a result, early diffusion steps are shared using the averaged embeddings, and the denoising steps gradually diverge to individual prompt embeddings, resulting in saved computation while maintaining high image generation quality.}
    \label{fig:tree-traversal}
    \vspace*{-8pt}
\end{figure*}

\begin{figure}
    \centering
    \includegraphics[width=\linewidth]{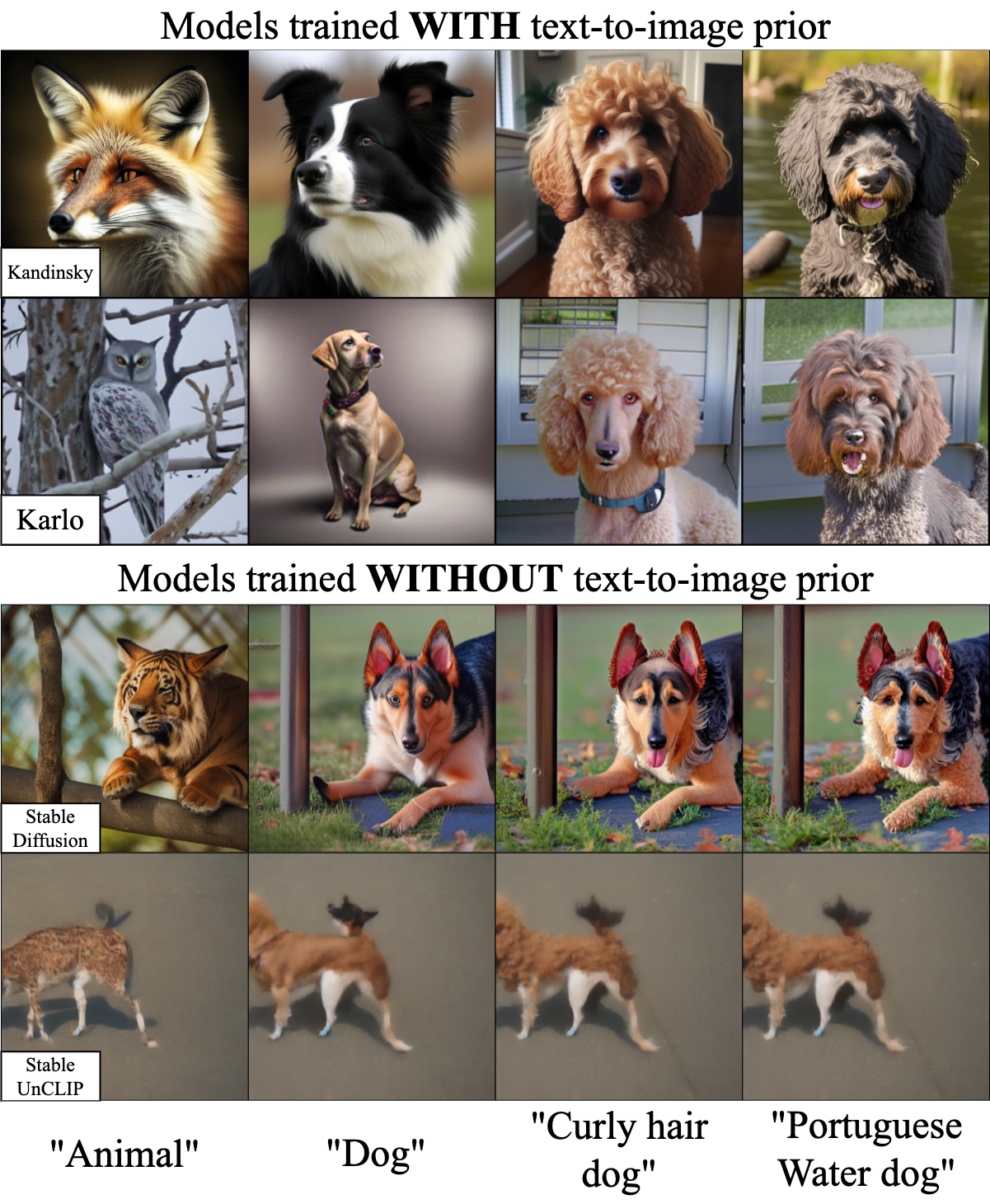}
    \caption{\textbf{Importance of text-to-image prior}. We share diffusion steps using a right-splitting binary tree as our hierarchy and the text prompts shown above for text-to-image diffusion models with various architectures. Kandinsky and Karlo are trained using an UnCLIP-style text-to-image prior. Stable diffusion does not use a text-to-image prior. While Stable UnCLIP does use a text-to-image prior during inference, the model was finetuned from Stable Diffusion and thus did not use it in training. Models trained with a text-to-image prior exhibit significantly higher quality and diversity with our approach than model trained without such prior.}
    \label{fig:unclip}
\end{figure}

\section{Method}
Our key insight is that two similar embeddings yield similar denoising updates during the early diffusion steps, introducing substantial redundancy.
Our idea is to initially share computation using this shared embedding $\tilde{y}$ for the first few denoising steps before progressively diverging into specialized prompt-specific embeddings. 
For two prompts, assuming we share the initial $\nicefrac{K}{2}$ steps, this simple strategy reduce the diffusion steps from $2K$ to $\frac{3K}{2}$.

For a large set of prompts, using a hierarchical tree structure maximizes computational reuse. This structure ensures that early diffusion computations -- corresponding to the widely shared visual features -- are performed only once at the root node, producing substantial computational savings. Subsequent computations then progressively branch out, specializing embeddings as the diffusion process advances toward fine-grained details. Notably, when structured as a balanced binary tree, as illustrated in \cref{fig:tree-traversal}, the computational complexity reduces from $\mathcal{O}(N)$ to $\mathcal{O}(\log N)$.
As the number of prompts increases, opportunities for computational sharing grow significantly, substantially reducing the total number of denoising steps required.

Next, we introduce a fully automated method to generalize this strategy for arbitrary prompt sets.

\paragraph{Constructing Embedding Tree.}

Given a set of $N$ text prompts encoded to latent codes $\mathcal{Y} = \{y_i\}_{i=1}^N$, our goal is to construct an embedding tree, exemplified by \cref{fig:tree-traversal}, where nodes near the root represent widely shared visual concepts, while nodes at lower levels specialize on more specific concepts shared only by a few prompts, culminating with leaf-level nodes corresponding to input prompts $y_i$. 

We construct the embedding tree using agglomerative clustering~\cite{Ward01031963}:
\begin{enumerate}
\item Initialize leaf nodes with embeddings corresponding directly to input prompts: ${c_1, \dots, c_N} = {\{y_1\}, \dots, \{y_N\}}$. %
\item Select two clusters with the smallest distance between them. Specifically, find clusters $(c_a, c_b)$ such that:
\begin{equation}
    (c_a, c_b) = \argmin_{c_a, c_b \in C, c_a \neq c_b} d(c_a, c_b),
\end{equation}
where the distance $d(c_a, c_b)$ is measured by cosine similarity between the mean embeddings of clusters:
\begin{equation}
    d(c_a, c_b) = 1 - \frac{\bar{e}_{c_a} \cdot \bar{e}_{c_b}}{\|\bar{e}_{c_a}\| \|\bar{e}_{c_b}\|}, \quad \text{where} \quad \bar{e}_c = \frac{1}{|c|}\sum_{y \in c} y.
\end{equation}

\item Merge clusters $(c_a, c_b)$ to form a new cluster $c_{ab} = c_a \cup c_b$ and assign it an embedding equal to the mean of its constituent embeddings $\bar{y}_{c_{ab}}$ computed as above.

\item Repeat the merging steps until a cluster containing all embeddings is added to the embedding tree.
\end{enumerate}
The compute spent in constructing the tree is negligible compared to the cost of the denoising diffusion process.
\paragraph{Adaptive Embedding Selection.}
Once the embedding tree is constructed, we define a function $f(y,k)$ to select the cluster whose centroid embedding will be used at a step $k$ to generate an image conditioned on $y$. At the final diffusion step ($k=K$), the diffusion step uses their leaf node embeddings.
At earlier steps, embeddings from higher-level ancestor nodes are selected.

We introduce a cluster heterogeneity score based on the distance between merged child clusters:
\begin{equation}
c^{\text{score}} = d(c_a, c_b),
\end{equation}
where $c_a, c_b$ are child clusters merged to form node $c$.
Lower heterogeneity indicate prompts within the cluster are semantically closer, suitable for guidance
during earlier diffusion steps.
Note that, by construction, the heterogeneity is always decreasing on any path from the root node to leaf.

During the diffusion process, we traverse the tree from the root to leaf nodes,
using the heterogeneity score to decide how many diffusion steps should be spent at each level of the tree.
More precisely, for each diffusion step $k$, we connect the $k$ to an acceptable level of heterogeneity through a function $\phi$, and select the child of the lowest node level that satisfies this constraint.
Formally, given a prompt embedding $y$, for each diffusion step $k$, we select the embedding from the node satisfying: 
\begin{equation}
\begin{split}
f(y,k) = \argmin_c c^\text{score}\text{, s.t.}\\
[\mathcal{P}\text{arent}(c)]^\text{score} \geq \phi(k)
\wedge y \in c,
\end{split}
\end{equation}
where $\phi(k)$ decreases linearly as the diffusion progresses:
\begin{equation}
    \phi(k) = \tau \cdot \left(1 - \frac{k}{K}\right).
\label{eq:phi}
\end{equation}
Here, $\tau$ is a hyperparameter ranging from 0 to $c^{\text{max}}$, the score of the root node, that controls how quickly embeddings become specialized. A lower $\tau$ forces embeddings to specialize earlier, resulting in lower savings and higher image quality, while a higher $\tau$ allows prompts to share embeddings longer, resulting in higher savings and lower image quality. In our experiments, high-quality generation and high savings can be achieved with a value of $\tau=1$.

We integrate this hierarchical embedding selection into standard diffusion inference.
The denoising process can be written as
\begin{align}
\label{eq:our_inference}
\x_T &\sim \mathcal{N}(\mathbf{0}, \mathbf{I}), \nonumber \\
\x_{t-1} &\sim \mathcal{N}\left(a_t \x_t - b_t \epsilon_\theta(\x_t; t, \bar{e}_{f(y,k)}), \sigma_t^2 \mathbf{I}\right).
\end{align}
Notice that while performing these steps to generate a set of images,
we can keep track of evaluations of $\epsilon_\theta(\x_t; t, \bar{e}_{f(y,k)})$
and make sure that they are not done more than once, which
significantly reduces computational cost without sacrificing image quality (see \cref{sec:experiments}).

\section{Experiments}
\label{sec:experiments}
We demonstrate our method's generation quality and compute savings on various benchmarks and applications. We present both quantitative and qualitative comparisons that show our method outperforms standard diffusion sampling, producing equal or higher quality generations using fewer total denoising steps for sets of images. We ablate key components of our method such as our semantic clustering and the condition modality of the underlying diffusion model. Finally, we show multiple applications of our method for real world personal and commercial tasks.

\subsection{Quantitative Evaluation}
\begin{figure}
    \centering
    \includegraphics[width=0.95\linewidth]{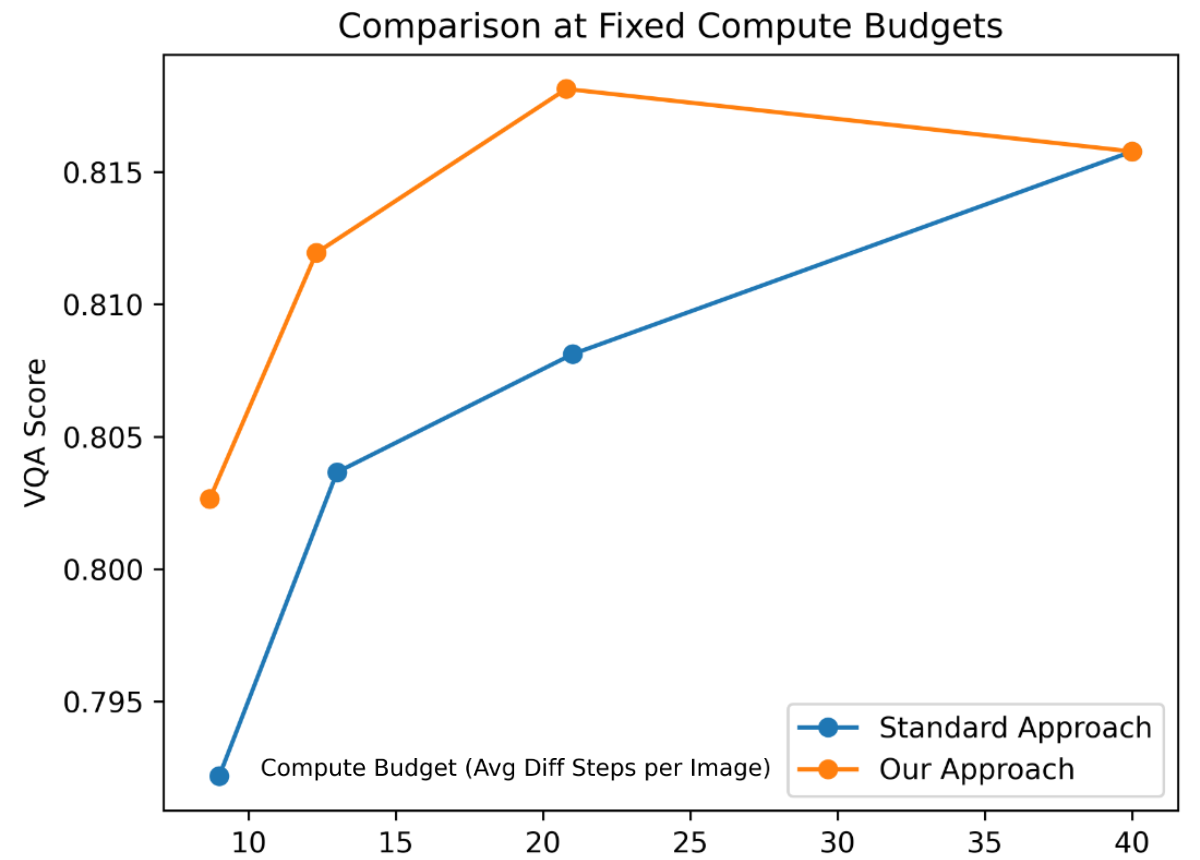}
    \vspace{-4pt}
    \caption{\textbf{Generation Quality at Fixed Compute Budgets}. We report VQA Score~\cite{lin2024evaluating} on the Prompt Template Dataset~\cite{promptartists} for both our method (orange) and the standard approach (blue) over various diffusion step budgets. Note that at a $40$-step budget, our approach is identical to the standard approach. For all other fixed compute budgets, our method achieves higher VQA Scores (a measure of generation quality) than the standard approach.}
    \label{fig:vqa-score}
\end{figure}

\begin{table}
\centering
\footnotesize
\begin{tabular}{@{ }l|ccc@{ }}
\toprule
Dataset & Num Images & Compute Saved $\uparrow$ & Win \% $\uparrow$\\
\midrule
GenAI Bench & 1600 & 50.43\% & 50.75\%\\
Prompt Template & 2000 & 69.25\% & 51.55\%\\
Style Variations & 100 & 73.93\% & 52.00\%\\
Animals & 500 & 62.80\% & 52.80\%\\
\bottomrule
\end{tabular}
\caption{\textbf{Compute savings using our approach.} We run our method on four diverse datasets ranging from most general (GenAI Bench~\cite{li2024genai}) to more structured (Style Variations) and compute the VQA Score~\cite{lin2024evaluating} for both our method and the standard approach. For each dataset, we report the number of images, the percentage of compute we save using our approach relative to standard denoising, and the percentage of images on which our approach achieves a higher VQA Score. From the win percentages, we can see that our method consistently produces comparable if not higher quality results to the standard approach, all while using significantly less compute (as evidenced by the compute saved percentages).}
\label{table:compute-savings}
\end{table}

\paragraph{Experimental and Evaluation Setup.}
Throughout this section we look to evaluate both generation quality and compute savings of our method. Since our method focuses on text-conditioned image generation, we use VQA Score~\cite{lin2024evaluating} as a measure of generation quality. VQA Score uses a visual-question-answering model to compute the probability of a ``Yes" answer to the question ``Does this figure show \{$p$\}" for a given image and its corresponding text prompt $p$. VQA Score significantly outperforms CLIP score in terms of alignment to human preference on text-to-image generation~\cite{li2024genai}. For our fixed compute experiments, we report this VQA score outright, while in experiments looking to establish comparative generation quality, we report the percentage of examples for which a given method achieves a higher VQA score. To evaluate compute savings, we report the percentage of steps saved relative to the standard approach. Unless specified otherwise, all experiments are run on the Kandinsky 2.2 model~\cite{razzhigaev2023kandinsky}.

\paragraph{Datasets.}
We evaluate our method on several diverse collections of prompts for text-driven image set generation varying in size, structure, and generality. GenAI Bench~\cite{li2024genai} is a text-to-image generative AI dataset containing $1600$ text prompts created by artists for the purpose of AI text-driven image generation. 
These prompts are diverse, challenging, and often require higher order reasoning such as logic and comparison.
We also use three custom datasets to understand how our approach performs on text prompts with more clear cut similarity due to their formulaic nature. The Prompt Template dataset uses the technique of prompt templates~\cite{promptartists} to generate $2000$ variations of the prompt ``A $<$art style$>$ painting of $<$subject$>$, $<$location$>$" by combinatorial replacement of each bracketed component. The Animals dataset is a set of $500$ prompts describing animals generated with an LLM. The Style Variations dataset is a set of $100$ prompts describing image variations to the prompt ``A child sitting on a swing, looking up at the stars, $<$style$>$" (see \cref{fig:image-styles}).

\paragraph{Generation Quality Under Fixed Compute Budget.} In \cref{fig:vqa-score},
we compare our method to standard diffusion using the Prompt Template dataset over multiple fixed compute budgets. At the full $40$ step budget, each generation is allowed its own $40$ steps and thus there is no sharing of compute leading our approach to become identical to standard diffusion. However, for fixed compute budgets less than this maximum $40$ step budget, our method obtains significantly higher VQA scores indicating superior generation quality. Notably, our approach with a compute budget of an average approximately $31$ steps per image produces higher VQA scores than the standard non-shared approach with a full $40$ step budget. This implies that our method is able to achiever higher quality generation by sharing compute than by using standard diffusion. We hypothesize that this is due to average text prompt conditions at early stages in the diffusion process producing a more stable denoising trajectory.

\paragraph{Compute savings under fixed generation quality.}
While our method outperforms standard diffusion at fixed compute budgets, it also enjoys significant compute savings for a fixed level of image quality. Specifically, we investigate how much compute can be saved with our approach to achieve similar quality as the $40$-step standard diffusion. We report both the percentage of average diffusion steps saved relative to the standard approach at 40 steps, as well as the percentage of images on which our approach achieves a higher VQA Score. From \cref{table:compute-savings}, we can observe that our method achieves comparable if not better quality on all four datasets while saving anywhere from $50\%$ to $74\%$ of the computational cost.

\subsection{Qualitative Comparisons}
\begin{figure*}
    \centering
    \vspace{2mm}
    \newcommand{\pl}{34}
    \newcommand{\pll}{-3}
    \begin{overpic}[width=\linewidth]{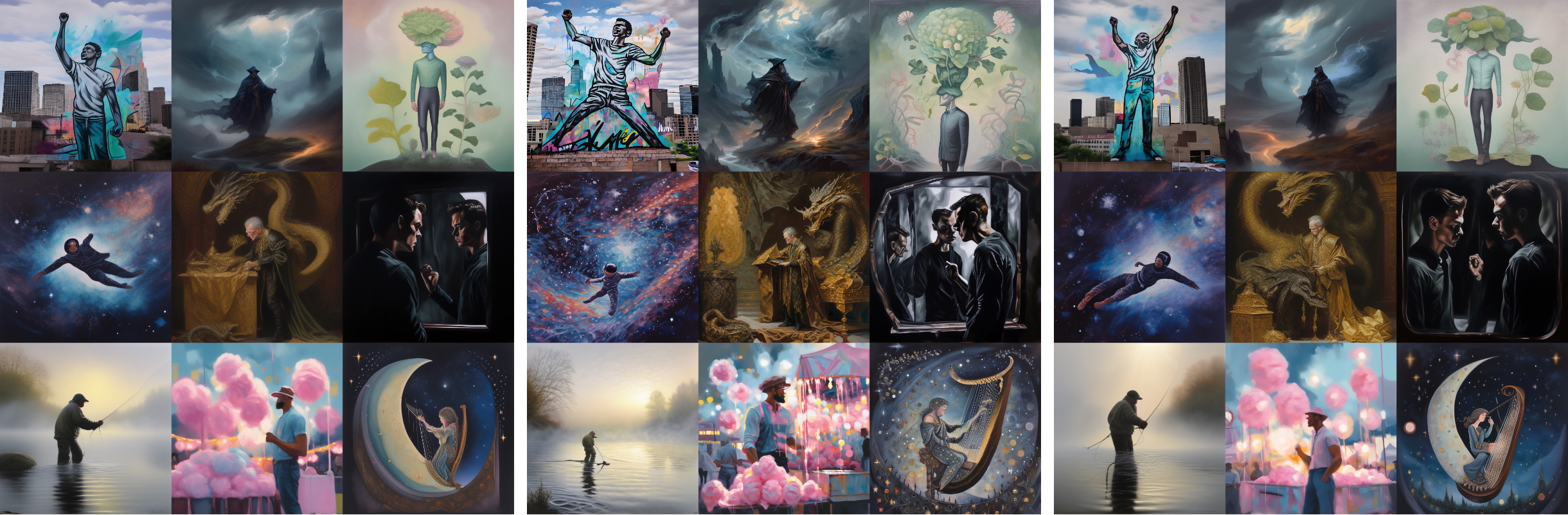}
    \put(9,  \pl){\textcolor{black}{Standard Diffusion}}
    \put(48,  \pl){\textcolor{black}{Ours}}
    \put(76,  \pl){\textcolor{black}{Standard Diffusion}}
    \put(6,  \pll){\textcolor{black}{Compute Budget: 18 Steps}}
    \put(38,  \pll){\textcolor{black}{Compute Budget: 17.8 Steps}}
    \put(72,  \pll){\textcolor{black}{Compute Budget: 40 Steps}}
    \end{overpic}
    \vspace{1mm}
    \caption{\textbf{Qualitative Comparison}. We compare our approach (middle) with standard diffusion for a fixed compute budget of $18$ steps (left) and a full compute budget of $40$ steps (right). For the fixed compute budget, our method produces higher quality images than the standard approach. When compared to the $40$ step compute budget, our approach uses significantly less steps while generating images of comparable if not higher quality.}
    \vspace{-3mm}
    \label{fig:qualitative-comparison}
\end{figure*}
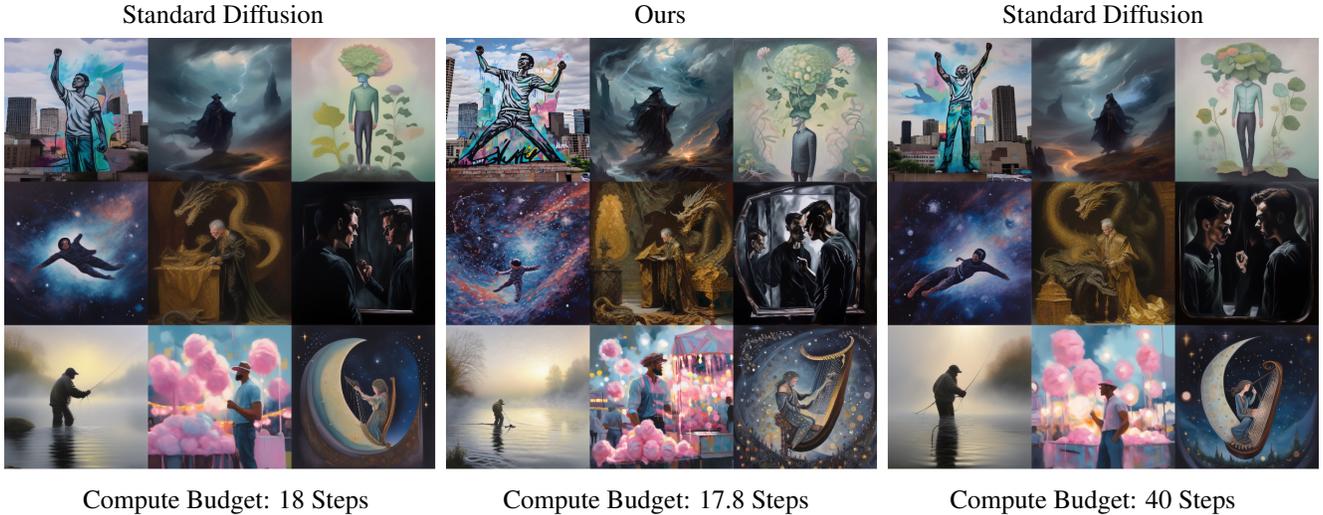

In addition the quantitative metrics presented above, we show qualitative comparisons of our approach with standard diffusion sampling for both a fixed compute budget of $18$ steps and a full $40$ step budget. In \cref{fig:qualitative-comparison}, we run these methods on a set of $500$ LLM-generated text prompts and display $9$ randomly selected images. Enforcing a compute budget of $18$ steps in standard diffusion produces over-smoothed images that lack fine-grained details. In contrast, our method is able to generate sharp features such as the wrinkles on the man's shirt (bottom row, middle column) or the sharp contours of the clouds (top row, middle column). When compared to the full $40$ step generation with standard diffusion, our method uses significantly less compute, while producing images of comparable detail and quality.

\subsection{Ablations}
\begin{table}
\centering
\footnotesize
\begin{tabular}{@{ }lcccc@{ }}
\toprule
&\multicolumn{4}{c}{Compute Saved $\uparrow$}\\
& GenAI & Prompt & Style &\\
Dataset & Bench & Template & Variations & Animals\\
\midrule
Random Clustering & 5.43\% & 5.48\% & 3.98\% & 4.89\%\\
Ours & \textbf{50.43\%} & \textbf{69.25\%} & \textbf{73.93\%} & \textbf{62.80\%}\\
\bottomrule
\end{tabular}
\caption{\textbf{Clustering Ablation.} We ablate the our semantic clustering algorithm by comparing to a baseline approach that clusters based on random encodings. Similar to Table 1,  we report the percentage of compute we save using our approach relative to standard denoising, both tuned to achieve quality similar to 40-step diffusion (measured by a VQA score of about 50\%). For all datasets, our clustering algorithm saves massively more compute.}
\label{table:clustering-ablation}
\end{table}

We run several ablations on key components of our method to demonstrate the effect of these design choices. To highlight the importance of our semantic clustering approach, we pass random embeddings to the hierarchical clustering algorithm and report the results in \cref{table:clustering-ablation}.
Since every cluster is more heterogeneous, the random baseline diverges very early to the leaf nodes which leads to little compute being saved.

We also evaluate our approach across various backbone models. As seen in \cref{fig:coarse-to-fine}, models trained with a text-to-image prior generate structure more evenly during the diffusion process than models that are not. Empirically, we observe that our approach performs much better on these models. In \cref{fig:unclip}, we run our approach on the set of four prompts ``Animal," ``Dog," ``Curly hair dog," and ``Portuguese water dog." Since the embeddings for ``Curly hair dog" and ``Portuguese water dog" are very similar, these two generations share most of the diffusion steps and only diverge on the final denoising stages to their respective prompts. In models trained with a text-to-image prior such as Kandinsky (first row) and Karlo (second row), this is not an issue as the structure is not fully set and fine-grained details can be generated in the remaining steps. Conversely, for models that were not trained with this prior such as Stable Diffusion (third row) and Stable UnCLIP (fourth row), there are not enough remaining steps to properly distinguish these two.

Finally, we investigate different values of $\tau$ for the mapping function $\phi$ of \cref{eq:phi}. Since the diffusion steps are evenly spaced out between $\phi(1) = \tau$ and $\phi(K)=0$, by increasing or decreasing $\tau$, we can control where on the hierarchy the diffusion steps are concentrated and thus the amount of steps shared by our approach. Using this, we can effectively control our compute budget. In \cref{fig:vqa-score}, we show our approach (orange) for $\tau$ values $[0, 0.5, 1, 1.5]$. While we leave $\tau$ as a hyperparameter users can tune to achieve a desired trade-off between savings and quality, we found that a value of $\tau=1.0$ worked well in most scenarios. We also investigated using non-linear mappings for $\phi$ such as a log scale, however in practice, the linear map worked best.

\subsection{Applications}
\begin{figure}
    \centering
    \newcommand{\pl}{-5}
    \newcommand{\pll}{-10}
    \begin{overpic}[width=\linewidth]{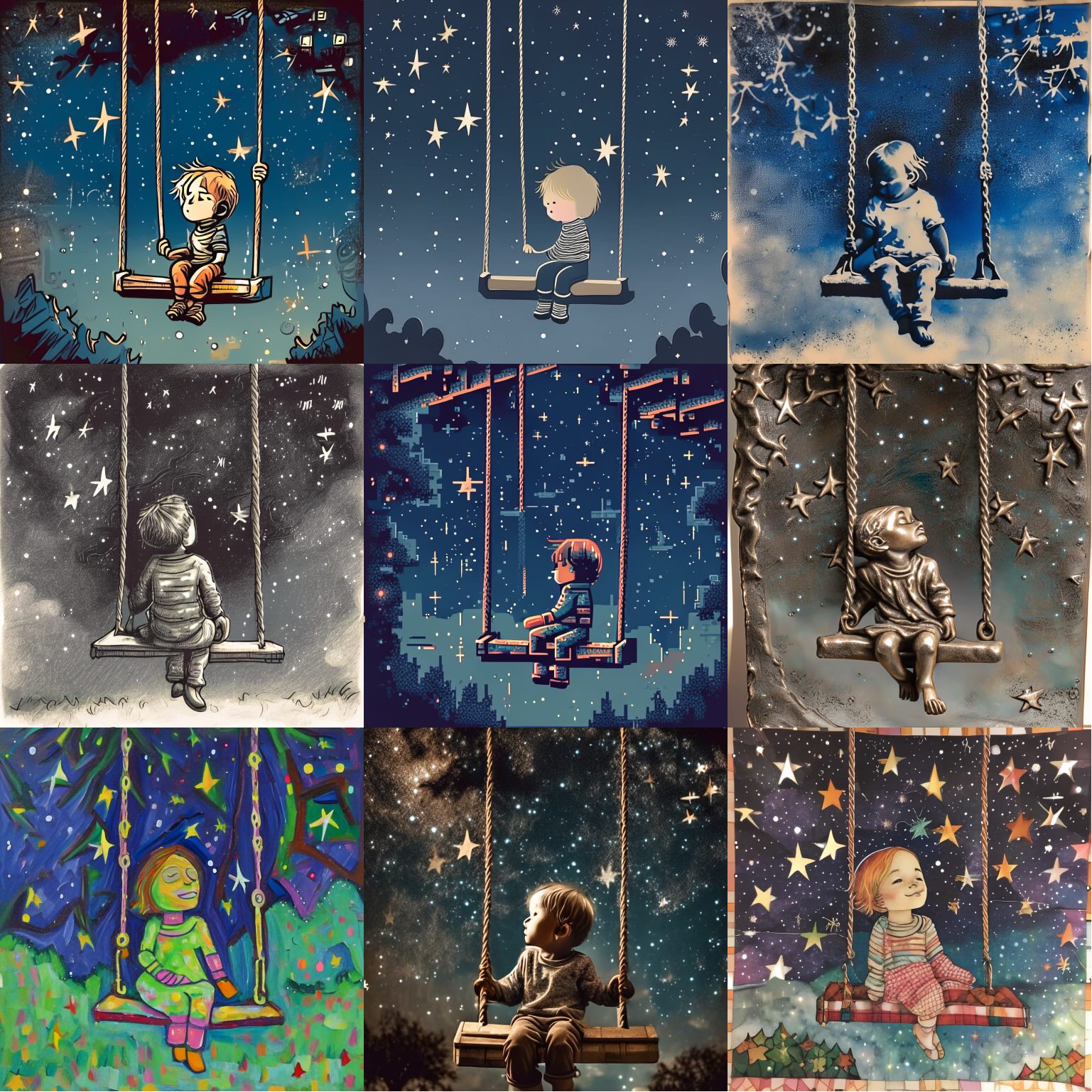}
    \footnotesize{
        \put(20,  \pl){\texttt{``A child sitting on a swing, }} 
        \put(16,  \pll){\texttt{looking up at the stars, $<$style$>$"}}
    }
    \end{overpic}
    \vspace{4mm}
    \caption{\textbf{Image Style Variations}. We show an application of our method for efficiently generating style variations on given input prompt. We show a subset of 100 generated images, saving 74\% of the equivalent compute for the standard approach.}
    \label{fig:image-styles}
\end{figure}

\begin{figure}
    \centering
    \includegraphics[width=\linewidth]{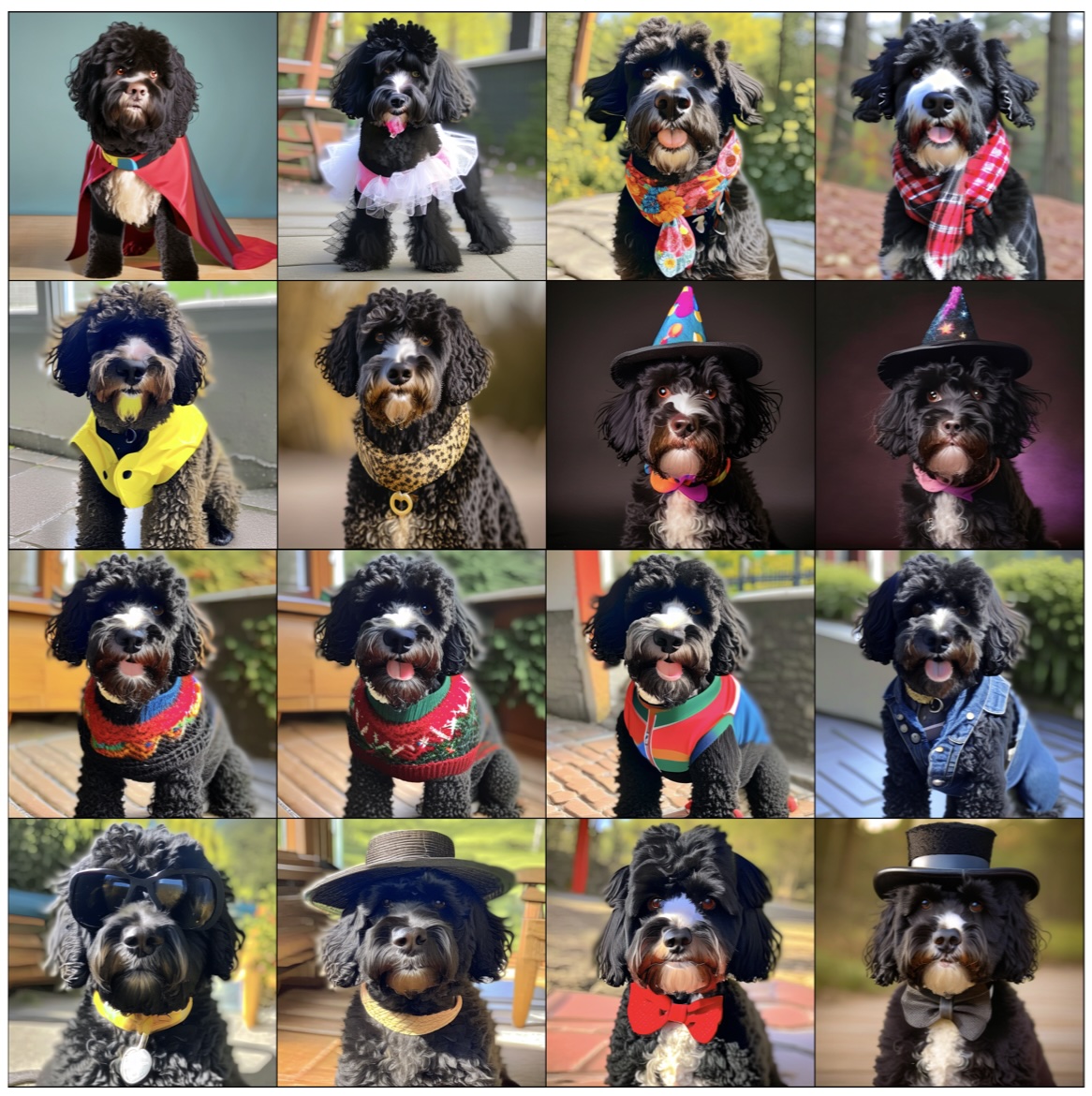}
    \vspace{-6mm}
    \caption{\textbf{Virtual try-on}. Our method generates a set of only 16 images, all containing the same subject, with various accessories, saving 65.3\% of the equivalent compute for the standard approach.}
    \label{fig:accessories}
\end{figure}

\begin{figure}
    \centering
    \newcommand{\pl}{-5}
    \newcommand{\pll}{-10}
    \begin{overpic}[width=\linewidth]{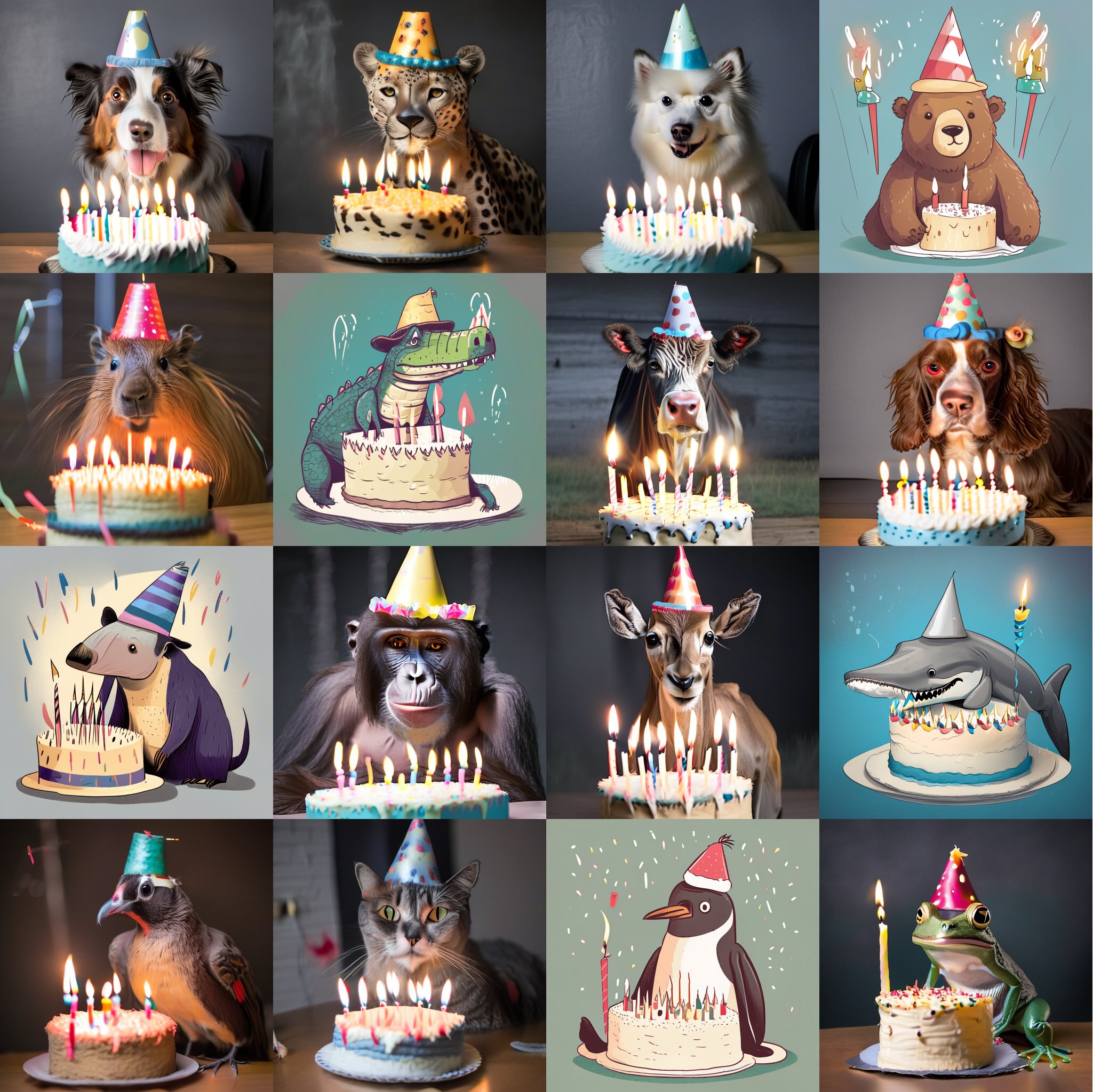}
    \footnotesize{
    \put(2,  \pl){\texttt{``An image of a $<$animal$>$ wearing a birthday hat}}
    \put(17,  \pll){\texttt{with a cake and birthday candles"}}
    }
    \end{overpic}
    \vspace{3mm}
    \caption{\textbf{Subject Variation}. We show an application of our method for efficiently generating subject variations on given input prompt. We show a subset of 500 generated images, saving 76\% of the equivalent compute for the standard approach.}
    \label{fig:subject-variation}
\end{figure}

In \cref{fig:image-styles,fig:accessories,fig:subject-variation}, we present three applications of our method to efficiently generate image variations using prompt templates, including stylistic and structural variations. Our approach achieves a compute saving of 74\% compared to the standard method when generating 100 style variations, 76\% for 500 subject variations, and 65.3\% for the 16 virtual try-on variations. Notably, the latter demonstrates significant computational efficiency even with relatively small image sets, provided they share common structural content. From a user perspective, efficiently generating image variations is helpful for iterating and co-creating with the diffusion model.

\subsection{Limitations and Future Work}
Our approach relies on factorizing denoising steps across similar samples, which introduces two primary limitations. First, our method is particularly suited to diffusion models where detail generation is evenly distributed throughout the diffusion process. As illustrated in \cref{fig:coarse-to-fine}, Stable Diffusion~\cite{rombach2022high} tends to establish structural information very early in sampling, significantly reducing the opportunities for factorizing denoising steps effectively. Second, our technique excels in scenarios where the image set is either large (and possibly diverse) or small but homogeneous; however, its performance diminishes in situations where the image set is small yet highly diverse. In these cases, factorization of computational effort becomes less beneficial, leading to reduced efficiency.

While our method is primarily designed for diffusion, we are interested in extending it to autoregressive models (AR). In particular, we anticipate that recent autoregressive models such as the VAR approach~\cite{tian2025visual}, which generate images in a coarse-to-fine manner via next-scale prediction, could also be adapted to benefit from sharing computation.

\section{Conclusion}

We introduced a method to reuse computational resources when generating multiple images using  diffusion models.
Our key insight is that, if images contain similar content, they can share early stages of the denoising process.
Leveraging this idea, we developed a technique that organizes target prompts into a hierarchical tree structure, where each node represents the average text embedding of its associated leaf prompts.
This structure enables effective computation factorization during image generation.
Through empirical analysis, we observed that diffusion models trained with text-to-image priors
are particularly suitable for computation factorization. For this category of models, our approach consistently achieves at least 50\% computational savings for diverse sets of prompts and even greater savings for more homogeneous sets,  while preserving image quality.
We demonstrated various practical applications, including generating image variations based on style and content.

We hope that this research will inspire further studies aimed at enhancing the efficiency of generative models, reducing environmental impact, and lowering carbon footprints.

\medskip
\noindent\textbf{Acknowledgments.} We thank Richard Zhang for insights on evaluation metrics and more generally, the members of Adobe Research and 3DL for their insightful feedback. This work was supported by Adobe Research and NSF grant 2140001.
{
    \small
    \bibliographystyle{ieeenat_fullname}
    \bibliography{main}
}

\clearpage
\setcounter{page}{1}
\maketitlesupplementary
\appendix

\section{Further Method Details}
We provide pseudocode for our hierarchical diffusion algorithm.
Our inputs are as follows: $\tau$ is a hyperparameter controlling the tradeoff between savings and quality, $K$ is the total number of diffusion steps, $C$ is our set of clusters obtained through agglomerative clustering, $C^{scores}$ are the corresponding $c^{score}$s for each cluster in $C$, and $Y$ is our set of all text prompts. The additional variables in line $18$ are standard diffusion variables: $a_k$ and $b_k$ are scheduler coefficients, $\epsilon_{\theta}$ is the diffusion model, and $\sigma_{k-1}$ is a time-step-dependent standard deviation. Note, this algorithm assumes that $\tau \leq max_c(c^{score})$ otherwise a dummy parent node is required above the root node with a corresponding $c^{score} > \tau$. Our method returns the image from step $K$ for each leaf cluster containing an individual prompt.
\begin{algorithm}
    \begin{algorithmic}[1]
        \State Inputs: $\tau, K, C, C^{\text{scores}}, Y$
        \State Let $\phi(k) = \tau \cdot \left(1 - \frac{k}{K}\right)$ assign scores to each diffusion step $k \in 1...K$ s.t. the scores of the $K$ steps are evenly spaced over the interval $[\tau, 0]$.
        \For{$k$ in $1...K$}:
            \State $C_{\text{k}} \gets []$
            \For{$y$ in $Y$}:
                \State $c' = \argmin_{c \in C} c^\text{score}$ s.t. $[\mathcal{P}\text{arent}(c)]^\text{score} \geq \phi(k)\wedge y \in c$
                \If{$c'$ not in $C_{\text{k}}$}
                    \State $C_{\text{k}}$.append($c'$)
                \EndIf
            \EndFor
            
            \For{$c$ in $C_{\text{k}}$}
                \If{k==1}
                    \State $\x^c_{k-1} \sim N(0, \textbf{I})$
                \Else
                    \State $\x^c_{k-1} \gets \x^{\mathcal{P}\text{arent}(c)}_{k-1}$
                \EndIf
                \State $\bar{y} = \text{mean}(y \text{ : } y \in c)$
                \State $\x^c_{k} \sim \mathcal{N}\left(a_k \x^c_{k-1} - b_k \epsilon_\theta(\x^c_{k-1}; k-1, \bar{y}), \sigma_{k-1}^2 \mathbf{I}\right)$
            \EndFor
        \EndFor
        \State\Return $\{\x^{\{y\}}_K : y \in Y\}$
    \end{algorithmic}
    \caption{Hierarchical Diffusion.}
    \label{alg:aes}
\end{algorithm}

\section{Additional Experiments}
\noindent \textbf{Experiments on Additional Models.}
\noindent Due to the coarse-to-fine generation properties exhibited by models conditioned on a text-image prior, our approach works best on the Kandinsky and Karlo models which are trained in this manner (See Fig. 4). However, we can still achieve savings using more standard models such as SD 1.5 and models using the more modern DiT architecture and Flow Matching scheduler such as FLUX.
While these models generate high frequency details earlier on in the diffusion process and thus leave less room for shared compute (see Fig. 2), we still achieve moderate savings over standard diffusion inference.
For comparable quality results (better VQA scores in $\approx50\%$ of the samples), we save up to $\textbf{28.25\%}$ and $\textbf{23.73\%}$ on compute when using SD 1.5 and FLUX, respectively.
We report results for both models on all four datasets in the table below.
\begin{table}[h]
\centering
\footnotesize
\begin{tabular}{@{ }cr|cccc@{ }}
\toprule
Model & Dataset & GenAI B. & Prompt T. & Style V. & Animals\\
\midrule
\multirow{2}{*}{SD 1.5} & Savings $\uparrow$ & 10.4\% & 28.25\% & 23.55\% & 11.70\%\\
& Win \% $\uparrow$ & 48\% & 50\% & 49\% & 50\%\\
\midrule
\multirow{2}{*}{FLUX} & Savings $\uparrow$ & 2.33\% & 3.61\% & 23.73\% & 6.28\%\\
& Win \% $\uparrow$ & 45\% & 45\% & 51\% & 48\%\\
\bottomrule
\end{tabular}
\vspace{-3mm}
\label{table:modern-savings}
\end{table}

\medskip
\noindent \textbf{Diversity of Generations.}
Examining sample generations from both our approach and standard diffusion in Figs. $1$ and $6$, we observe that our generations qualitatively display comparable diversity to the results from standard diffusion. Since our approach leverages similarities across prompts to save compute in the generation process, when used at small scales, our generations can exhibit high similarity between examples (see Fig. $8$).
However, as we increase the size of our datasets such as with the GenAI Bench or Prompt Templates datasets, our diversity significantly increases. To quantitatively evaluate generation diversity, we randomly sample $100$ images from each generated dataset and compute the CLIP cosine similarity between all possible pairs. The reported result is the mean similarity. On larger datasets (GenAI Bench, Prompt Templates, and Animals which all have $\geq 500$ examples), our method achieves comparable diversity to the standard approach. While our results on the Style Variations dataset are less diverse, this is expected since the dataset contains only $100$ examples and the highly structured and similar prompts naturally lend themselves to less diverse generations.
\begin{table}[h]
\centering
\footnotesize
\begin{tabular}{@{ }r|cccc@{ }}
\toprule
Dataset & GenAI B. & Prompt T. & Style V. & Animals\\
\midrule
Std. CLIP Diversity $\downarrow$ & 0.6082 & 0.6508 & \textbf{0.8071} & 0.6652\\
Our CLIP Diversity $\downarrow$ & 0.6027 & 0.6493 & 0.8374 & 0.6552\\
\bottomrule
\end{tabular}
\vspace{-4mm}
\label{table:diversity-metric}
\end{table}

\begin{figure*}[t]
    \centering
    \includegraphics[width=\linewidth]{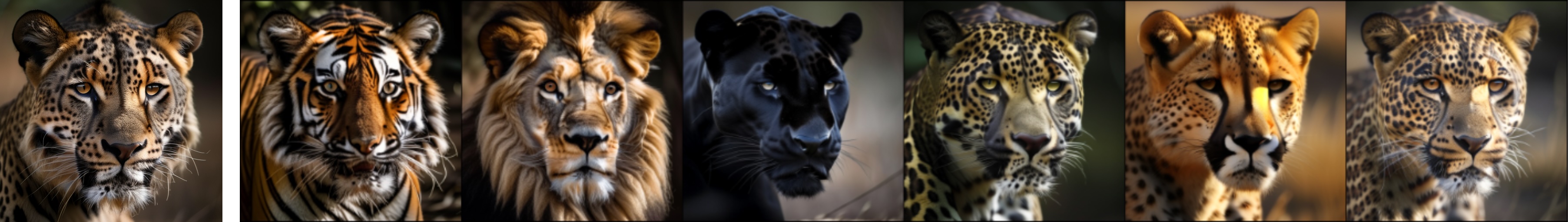}
    \vspace{-6mm}
    \caption{De-noised mean embedding (left) and its children using the Kandinsky model.}
    \vspace{-4mm}
    \label{fig:mean-embeddings}
\end{figure*}

\medskip
\noindent \textbf{Visualization of Mean Embedding.}
We analyze the mean cluster embedding by using it to guide image generation. As we can see in \cref{fig:mean-embeddings}, even though this embedding (left) is not associated with any particular prompt, it yields an image corresponding to a reasonable visual mixture of its children. In practice, this ``mean image'' is never generated (fully denoised) -- we show it here as a tool to visualize the semantics of that embedding.

\end{document}